\documentclass[sigconf]{acmart}
\AtBeginDocument{%
  \providecommand\BibTeX{{%
    \normalfont B\kern-0.5em{\scshape i\kern-0.25em b}\kern-0.8em\TeX}}}

\usepackage{microtype}
\usepackage{graphicx}
\usepackage{booktabs} 
\usepackage{subcaption}
\usepackage{amsmath}
 
\usepackage{amssymb}
\usepackage{mathtools}
\usepackage{amsthm}
\usepackage[capitalize,noabbrev]{cleveref}
\usepackage[textsize=tiny]{todonotes}
\usepackage{diagbox}
\usepackage{graphics}
\usepackage{xspace}
\usepackage{siunitx}
\usepackage{wrapfig}
\usepackage{color}
\usepackage{multirow}
\usepackage{float}
\usepackage{amsfonts}
\usepackage{bm}
\usepackage{array}
\usepackage{colortbl}
\usepackage{rotating}
\usepackage{enumitem}
\usepackage{listings}
\usepackage{pifont}
\usepackage{makecell}
\usepackage{arydshln}

\definecolor{mygray}{gray}{.92}
\usepackage{algorithm}
\usepackage{algorithmic}

\setcopyright{acmcopyright}
\copyrightyear{2023}
\acmYear{2023}
\setcopyright{rightsretained}

\acmConference[UbiComp/ISWC '23 Companion] {Adjunct Proceedings of the 2023 International Joint Conference on Pervasive and Ubiquitous Computing \& the 2023 International Symposium on Wearable Computers}{October 8--12}{Cancun, Quintana Roo, Mexico}

\acmBooktitle{Adjunct Proceedings of the 2023 International Joint Conference on Pervasive and Ubiquitous Computing \& the 2023 International Symposium on Wearable Computers (UbiComp/ISWC '23 Companion), October 8--12, 2023, Cancun, Quintana Roo, Mexico}

\acmPrice{15.00}
\acmDOI{10.1145/3594739.3612914}
\acmISBN{979-8-4007-0200-6/23/10}

\makeatletter
\gdef\@copyrightpermission{
 \begin{minipage}{0.3\columnwidth}
  \href{https://creativecommons.org/licenses/by/4.0/}{\includegraphics[width=0.90\textwidth]{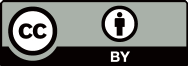}}
 \end{minipage}\hfill
 \begin{minipage}{0.7\columnwidth}
  \href{https://creativecommons.org/licenses/by/4.0/}{This work is licensed under a Creative Commons Attribution International 4.0 License.}
 \end{minipage}
 \vspace{5pt}
}
\makeatother
\begin{document}
\sloppy
\makeatletter
\newcommand{\thickhline}{%
	\noalign {\ifnum 0=`}\fi \hrule height 1pt
	\futurelet \reserved@a \@xhline
}
\makeatother

\title[Inclusive Data Representation in Federated Learning]{Inclusive Data Representation in Federated Learning: A Novel Approach Integrating Textual and Visual Prompt}

\author{Zihao Zhao}
\affiliation{%
  \institution{Tsinghua-Berkeley Shenzhen Institute,}
  \institution{Tsinghua University,}
  \city{Shenzhen}
  \state{Guangdong}
  \country{China}
}
\email{zhao-zh21@mails.tsinghua.edu.cn}

\author{Zhenpeng Shi}
\affiliation{%
  \institution{Tsinghua-Berkeley Shenzhen Institute,}
  \institution{Tsinghua University,}
  \city{Shenzhen}
  \state{Guangdong}
  \country{China}
}
\email{shizp22@mails.tsinghua.edu.cn}

\author{Yang Liu}
\affiliation{%
  \institution{Institute for AI Industry Research,}
  \institution{Tsinghua University,}
  \state{Beijing}
  \country{China}\\
  \institution{Shanghai Artificial Intelligence Laboratory,}
  \state{Shanghai}
  \country{China}
}
\email{liuy03@air.tsinghua.edu.cn}

\author{Wenbo Ding}
\authornote{Corresponding author}
\affiliation{%
  \institution{Tsinghua-Berkeley Shenzhen Institute,}
  \institution{Tsinghua University,}
  \city{Shenzhen}
  \state{Guangdong}
  \country{China}\\
  \institution{Shanghai Artificial Intelligence Laboratory,}
  \state{Shanghai}
  \country{China}
}
\email{ding.wenbo@sz.tsinghua.edu.cn}

\settopmatter{printacmref=true}


\begin{abstract}
Federated Learning (FL) is often impeded by communication overhead issues. Prompt tuning, as a potential solution, has been introduced to only adjust a few trainable parameters rather than the whole model. However, current single-modality prompt tuning approaches fail to comprehensively portray local clients' data. To overcome this limitation, we present Twin Prompt Federated learning (TPFL), a pioneering solution that integrates both visual and textual modalities, ensuring a more holistic representation of local clients' data characteristics. Furthermore, in order to tackle the data heterogeneity issues, we introduce the Augmented TPFL (ATPFL) employing the contrastive learning to TPFL, which not only enhances the global knowledge acquisition of client models but also fosters the development of robust, compact models. The effectiveness of TPFL and ATPFL is substantiated by our extensive evaluations, consistently showing superior performance compared to all baselines.
\end{abstract}





\maketitle

\section{Introduction}
The emergence of distributed learning systems has provided considerable advantages across a wide range of domains. Nonetheless, growing privacy concerns about distributed learning have necessitated the advent of \textit{Federated Learning} (FL) \cite{mcmahan2017communication, bonawitz2022federated}, a framework expressly developed to protect participants' private information. In FL, instead of uploading their private data, local clients share their local model weights with a central server during each communication round. The server aggregates these models and circulates them back to the local clients, thereby accomplishing the goal of information consolidation.

Recently, FL has confronted a wealth of challenges, including significant communication overheads \cite{ZHAO2023, 10185584, zhao2023aquila} and data heterogeneity \cite{kairouz2021advances}. A variety of recent research initiatives have sought to tackle these obstacles. Specifically, some have proposed innovative efficient encoding and model compression algorithms to reduce the communication cost, such as quantization to a continuous range of values into a finite set and sparsification \cite{sattler2019robust} to clip the full gradient into a sparse one, as well as intelligent scheduling of client participation \cite{nishio2019client} during the training process. Moreover, some incorporate the original FL framework with an additional step of knowledge distilling \cite{lin2020ensemble} to contract larger models into smaller ones, thereby enhancing the robustness of the global model. 

Despite these strategies, certain inherent limitations persist. Primarily, they require a substantial volume of labeled training samples, which may be unavailable to many clients in the FL environment, hindering effective training and resulting in model overfitting \cite{jin2020towards}. In addition, notwithstanding the communication costs reduction achieved by these efficient methods, most IoT devices such as smart home devices or industrial sensors, cannot accommodate large backbone model training due to their limited processing powers \cite{imteaj2021survey}, infinitesimal memory, and energy constraints. To illustrate, training a ResNet-50 model \cite{he2016deep} involves intensive computation and storage memory. It has approximately 25 million weight parameters and computes 16 million activations in the forward pass. Even after the communication-efficient algorithm to weights and activations, the total storage needed for saving ResNet-50's intermediate gradient results is over 7.5 GB for a mini-batch of 32 on a high-performance GPU. Given the hardware constraints of typical IoT devices, it is clear that they would struggle to support such intensive computations and memory requirements.

To resolve these problems, current research is leaning towards prompt tuning \cite{lester2021power}. Unlike conventional fine-tuning methods in FL that tune and aggregate full model parameters, applying prompt learning in FL only adjusts soft prompts for corresponding downstream tasks, while keeping large backbone models static to diminish both the communication and computation costs. Back to the ResNet-50 case, prompt tuning could save gradient results to just a handful of MB, drastically decreasing the communication overhead. However, most existing work only considers a single modality, failing to represent the local clients comprehensively. For instance, \citet{guo2022promptfl} exclusively employs textual soft prompts to depict the local clients without taking the visual knowledge into consideration; yet, \citet{feng2023learning} leverages continuous visual prompts to capture the image data information, disregarding text knowledge. In contrast, our work proposes Twin Prompt Federated learning (TPFL), a method resorting to both visual and textual modalities for a more comprehensive representation of the local clients' data characteristics. First off, we find that merely combining two modalities overlooks the potential for a unified approach. As such, we devise Augmented TPFL (ATPFL) to fuse the contrastive learning approach into the prompt tuning, facilitating the acquisition of global knowledge by client models. To the best of our knowledge, ATPFL is the first to integrate both textual and visual modalities within the context of FL and use contrastive learning to connect them. The contributions of this paper are threefold:

\begin{itemize}
    \item We present an innovative FL framework named ATPFL, that merges both visual and textual modalities for an improved representation of local clients' data characteristics, surpassing existing work's performance that only considers a single modality.
    \item The incorporation of contrastive learning to prompt tuning, enabling clients to acquire more global knowledge and improving on the direct combination of modalities that may overlook the potential for a unified approach. This is the first work to integrate two modalities within the context of FL and to utilize contrastive learning for their integration.
    \item Extensive evaluations have been conducted to ascertain the effectiveness of TPFL and ATPFL. The results demonstrate that ATPFL outperforms all the baselines.
\end{itemize}

\begin{figure*}
    \centering
    \includegraphics[width=0.94\textwidth]{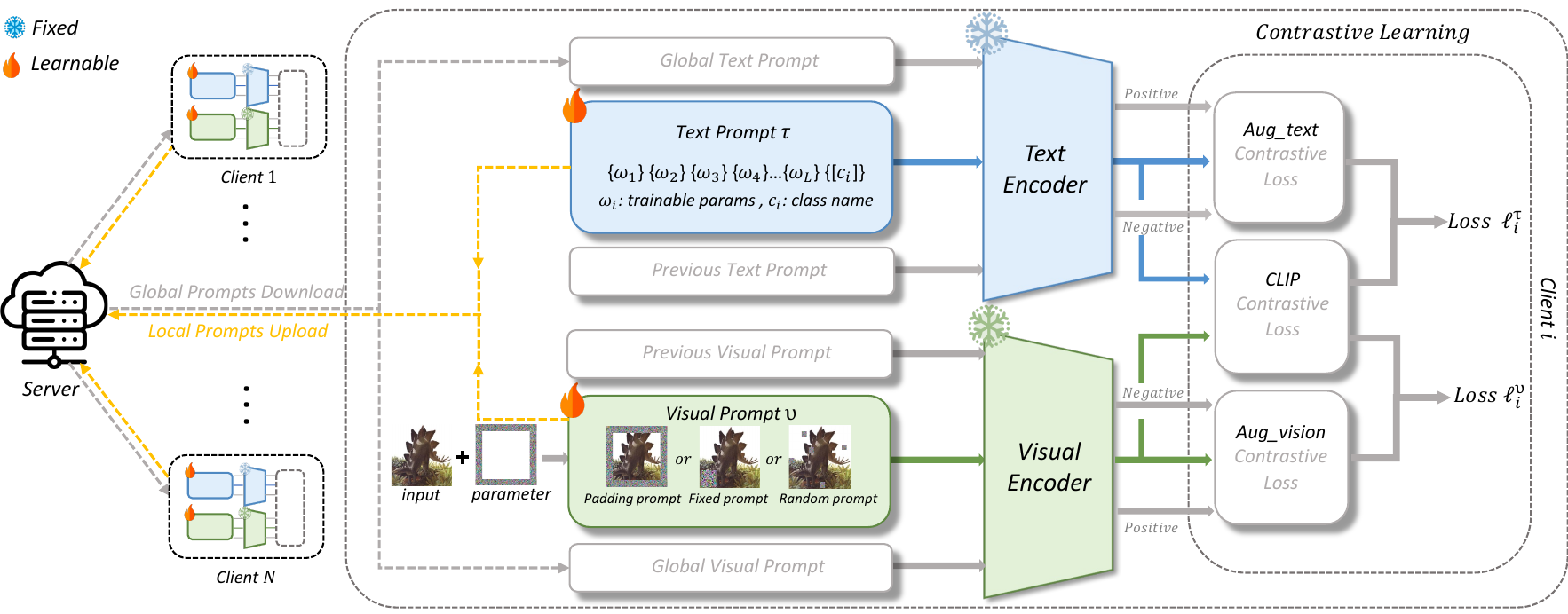}
    \vspace{-10pt}
    \caption{This figure illustrates the pipeline of ATPFL with contrastive learning. 
    In local training, the current prompt, previous prompt, and received global prompt are passed to each modality encoder. After the encoding, two types of contrastive learning are performed. Text contrastive loss and Visual contrastive loss use the feature extracted from the global prompt as positive contrast and the feature extracted from the previous prompt as negative contrast. CLIP contrastive loss is computed with the test prompt feature and the visual prompt feature. 
    }
    \label{framework}
\end{figure*}
\vspace{-10pt}

\section{Related Works}

\subsection{Communication Efficiency}
Communication efficiency has always been a critical challenge in the FL field. Different lines of research have been  investigated to tackle this challenge. 
Firstly, quantization\cite{gray1998quantization} methods are used to represent the full model parameters with lower bits. This technique involves converting the high-precision floating-point values of the model parameters into lower-precision values. For example, stochastic quantization\cite{alistarh2017qsgd} adaptively adjusts the quantization level in a stochastic manner. 
Secondly, sparsification methods improve communication efficiency by directly reducing the number of model parameters to be sent. More specifically, the sparsification method selects an important subset of model parameters and sets other insignificant parameters to zero before sending them to the global server. Top-k sparsification and rank-k sparsification are common sparsification methods\cite{eghlidi2020sparse}. Han et al.\cite{han2020adaptive}proposed to adaptively change the sparsification level to minimize overall training time. Shi et al.\cite{shi2019distributed} introduced global-k sparsification to compress the down streaming communication from the server to the clients.
Thirdly, knowledge distillation is also investigated to alleviate communication overhead\cite{li2019fedmd}. Knowledge distillation methods transfer knowledge from a larger teacher model to a smaller student model. Examples of knowledge-distillation-based federated learning are FedMD\cite{li2019fedmd}, FedDF\cite{sattler2020communication}, etc. However, all the aforementioned strategies have a high resource requirement and can hardly be implemented in IoT devices due to their limited hardware restrictions.

\subsection{Prompt Tuning}
\citet{houlsby2019parameter} proposed parameter-efficient transfer learning with adapter modules.  
\citet{liu2021p} showed that prompt-tuning can match the performance of fine-tuning with only 0.1\% - 3\% tuned parameters in the context of Natural Language Understanding.
\citet{li2021prefix} applied prefix-tuning to GPT-2 and BART for downstream tasks and shows that prefix-tuning can outperform fine-tuning in low-data settings. 
\citet{guo2022promptfl} proposed a federated learning framework for prompt-tuning called PromptFL.The PromptFL framework leverages the power of federated learning, which allows training prompts on decentralized data across multiple devices. In this work, only one modality text prompt is used and the result shows that federated prompt tuning achieved better performance compared to fine-tuning FL in many IID and non-IID settings.
Nonetheless, the existing research primarily focuses on a single modality, constraining their capability to obtain more information of local clients. In this paper, we present to employ both textual and visual representations to comprehensively characterize the local client.

\section{Methodology}
This section begins by outlining the basic structure of FL. Subsequently, we introduce the TPFL which considers both visual and textual information. Despite showing improvements, TPFL has certain inherent limitations. Therefore, we propose ATPFL to address these shortcomings and achieve superior performance.

\subsection{Problem Statement}
In the general FL setting, the entail system envelops $M$ clients, while, in every round, $K$ clients will actively participate, each possessing a unique local dataset. Each local dataset on client $k$ consists of $n_k$ samples, with each sample representing a pair, $(x_i^k, y_i^k)$, of a data feature $x$ and its corresponding target label $y$. The primary objective of FL is to construct a global model parameter vector $w$ that minimizes the mean loss across all local datasets, as demonstrated in the following optimization problem:
\begin{equation}
w=\arg \min _w \frac{1}{K} \sum_{i=1}^K \frac{1}{n_i} \sum_{n=1}^{n_i} \mathcal{L}\left(w ; x_n^i, y_n^i\right),
\end{equation}
where $w$ denotes the weights of the prediction model, $\mathcal{L}$ is the loss function.

\subsection{Twin Prompt Federated Learning (TPFL)}
As aforementioned CoOp \cite{zhou2022conditional} resorts to a series of continuous learnable parameters as the textual prompts, replacing the manually-designed constant ones. The textual prompt can be denoted as $\boldsymbol{\tau}_{c_i}=\left\{\boldsymbol{\omega}_1, \boldsymbol{\omega}_2, \ldots, \boldsymbol{c}_i, \ldots, \boldsymbol{\omega}_L\right\}$, where $\boldsymbol{c}_i$ signifies the word embedding of the $i^{th}$ image class names, $\boldsymbol{\omega}$ is a collection of learnable vectors, denoted as $\{\boldsymbol{\omega}_i|_{i=1} ^L\}$, and $L$ symbolizes the length of context words. Importantly, the position of $c_i$ can be placed anywhere between $(1, L + 1)$. In the training process, the textual prompt will be fed into a text encoder $g(\cdot)$, obtaining the textual feature as $z_{\tau_{c_i}}=g(\boldsymbol{\tau}_{c_i})$. Similarly, the visual feature $z_{\upsilon}=f(\boldsymbol{x})$ is calculated by visual encoder $f(\cdot)$. The final prediction probability is computed by the negative log likelihood matching score:
\begin{equation}
\vspace{-3pt}
    \ell_{con} (z_{\tau_{c_i}}, z_{\upsilon}) = -\log p(y=c_i \mid \boldsymbol{x})=-\log\frac{\exp \big(\operatorname{sim}\big(z_{\tau_{c_i}}, z_{\upsilon}\big) / \Gamma\big)}{\sum_{j} \exp \big(\operatorname{sim}\big(z_{\tau_{c_i}}, z_{\upsilon}\big) / \Gamma\big)},
    \label{contrastive}
\end{equation}
where $\operatorname{sim}(\cdot, \cdot)$ function represents the similarity function, $\Gamma \in \mathbb{R}$ is the temperature factor to control the overall distribution of the similarity between the embedding of the visual feature and test feature. 

Different from the previous work, which solely obtains a single modal to represent a local client, our study introduces TPFL  to resort to two different modalities, vision and text, to enhance the generalization capability and resilience of the global model. More specifically, instead of relying on a constant input visual feature $x$, we incorporate an additional trainable visual prompt $\upsilon$ as an extended representation for the local data characterization and conduct $x + \upsilon$ to get the final input feature. As illustrated in \cref{framework}, three templates of the visual prompt are employed: the padding, random patch, and fixed patch patterns, each contributing to varying model performances. After acquiring both the textual and visual prompts, each local client transmits them to the central server. The server then aggregated the received prompts, in light of the number of their training samples: 
\begin{equation}
    \boldsymbol{\tau}_g \leftarrow \sum_{i=1}^K \frac{n_i}{\sum_{j=1}^{K} n_j}  \boldsymbol{\tau}_i, \quad \boldsymbol{\upsilon}_g \leftarrow \sum_{i=1}^K \frac{n_i}{\sum_{j=1}^{K} n_j}  \boldsymbol{\upsilon}_i.
    \label{global_avg}
\end{equation}
However, the naive aggregation of the uploaded model weights may invite certain problems. To begin with, in practical scenarios, the data distribution across multiple clients may not be independently and identically distributed (IID). In other words, different clients can host data with significantly divergent statistical characteristics. The direct averaging of models struggles to effectively amalgamate local models originating from these devices, owing to this non-IID data distribution, and as a result, the performance of the global model suffers. Moreover, data volume can significantly vary across devices, with certain scenarios providing only a sparse dataset (only a few data points are available). Conventional FL aggregation might lack the robustness required to manage these few-shot learning scenarios, thereby complicating the process of discerning meaningful patterns from such limited data.

\subsection{Augmented TPFL (ATPFL)}

To address these aforementioned challenges, we propose the incorporation of a contrastive learning strategy, thus fortifying the robustness of FL. Specifically, we utilize the InfoNCE loss function \cite{oord2018representation} to encourage the output distributions of both the local visual and textual prompts to align closely with the output distribution of the global model. This methodology fosters a better comprehension of the global model by the local client, consequently mitigating the adverse effects of non-IID data. The key insight fueling this strategy is that contrastive learning facilitates the distinction between similar and dissimilar data points. It mitigates the discrepancies among local models caused by non-IID data through the learning of invariant features, making local models more amenable to aggregation at the global level. The contrastive (InfoNCE) loss functions for both textual and visual prompts are formulated in \eqref{con_text}:
\begin{multline}
\ell_{con\_aug}(z^{t+1}, z^t_g, z^t)=\\
-\log \frac{\exp \big(\operatorname{sim}\big(z^{t+1}, z^t_g\big) / \Gamma\big)}{\exp \big(\operatorname{sim}\big(z^{t+1}, z^t_g\big) / \Gamma\big)+\exp \left(\operatorname{sim}\left(z^{t+1}, z^t\right) / \Gamma\right)},
\label{con_text}
\end{multline}
where $z^{t + 1}$ and $z^{t}$ refer to the embedding of local textual or visual prompts at step $t+1$ and $t$, respectively, and $z_{g}^t$ represents the global textual and visual prompts.
After attaining the contrastive loss, the overall loss of the trainable prompts can be calculated by:
\begin{multline}
    \mathcal{L} (z_i^{t+1}, z^t_g, z_i^t) = \ell_{con} (z_{i, \tau}^{t+1}, z_{i, \upsilon}^{t+1}) \\
    + \mu (\ell_{con\_aug} (z_{i, \tau}^{t+1}, z_{g, \tau}^t, z_{i, \tau}^t) + \ell_{con\_aug} (z_{i, \upsilon}^{t+1}, z_{g, \upsilon}^t, z_{i, \upsilon}^t)),
\label{loss_vis}
\end{multline}
where $\ell_{con}$ denotes the contrastive loss formulated in \eqref{contrastive}, $z_{i, \tau}$ ($z_{g, \tau}$) and $z_{i, \upsilon}$ ($z_{g, \upsilon}$) denote the embedding of the local client $i$'s (global) textual or visual prompts, respectively, and $\mu$ represents a tuning factor to control the influence of textual augmented loss $\ell_{con\_aug} (z_{i, \tau}^{t+1}, z_{g, \tau}^t, z_{i, \tau}^t)$ and visual augmented loss $\ell_{con\_aug} (z_{i, \upsilon}^{t+1}, z_{g, \upsilon}^t, z_{i, \upsilon}^t)$. The overall training process of ATPFL is shown in \cref{algo_framework}.
\begin{algorithm}[tb]
  \caption{ATPFL}
  \label{algo_framework}
\begin{algorithmic}
  \STATE {\bfseries Input:} The entire $K$ clients are indexed by $i \in \{1, 2, \ldots, K\}$; $T_g$ and $T_{loc}$ is the number of global epochs and local epochs, respectively, and $\alpha$ is the learning rate.
     \STATE {\bfseries Server executes:} 
        \begin{ALC@g}
        \STATE Initialize $\boldsymbol{\tau}_g^{0}, \boldsymbol{\upsilon}_g^{0}$ 
         \qquad \FOR{each round $t = 1, 2, \ldots, T_g$}
        
            \FOR{each client $i$ {\bfseries in parallel}}
                \STATE$\boldsymbol{\tau}_i^{t + 1}, \boldsymbol{\upsilon}_i^{t + 1} \leftarrow ClientUpdate(i, \boldsymbol{\tau}_g^{t}, \boldsymbol{\upsilon}_g^{t})$
            \ENDFOR
            \STATE Aggregate the global prompts $\boldsymbol{\tau}_g^{t + 1}, \boldsymbol{\upsilon}_g^{t + 1}$ by \eqref{global_avg}
        \ENDFOR
         \end{ALC@g}
     
     \STATE {\bfseries ClientUpdate($i, \boldsymbol{\tau}_g^{t}, \boldsymbol{\upsilon}_g^{t}$):}

    \FOR{each local epoch from $1$ to $T_{loc}$}
        \STATE Calculate the logits and loss for textual prompt and visual prompt by \eqref{loss_vis}
        \STATE Update the local textual and visual prompts by gradient descent: 
        $\boldsymbol{\tau}_i^{t + 1} \leftarrow \boldsymbol{\tau}_g^{t} - \alpha \nabla \mathcal{L}_{\boldsymbol{\tau}_i^t}, \ \boldsymbol{\upsilon}_i^{t + 1} \leftarrow \boldsymbol{\upsilon}_g^{t} - \alpha \nabla \mathcal{L}_{\boldsymbol{\upsilon}_i^t}$
    \ENDFOR
    \STATE Return $\boldsymbol{\tau}_i^{t + 1}, \boldsymbol{\upsilon}_i^{t + 1}$ to the server
\end{algorithmic}
\end{algorithm}

\begin{table*}[t]
    \caption{Test Accuracy (\%) Results for ViT model on 7 datasets with 5 different seeds.}
    \vspace{-8pt}
    \centering
    \resizebox{0.99\textwidth}{!}{
        \setlength\tabcolsep{5pt}
        \renewcommand\arraystretch{1.02}
        \begin{tabular}{c||c c c c c c c }
            \thickhline
            \rowcolor{mygray}
            Algorithm (ViT) & Caltect-101	& Flowers-102 & Oxford-Pets & DTD & EuroSAT & Stanford Car & UCF-101\\
            \hline
            \hline
            Local Training &
            86.9$_{\textcolor{gray}{\pm0.03}}$ & 58.7$_{\textcolor{gray}{\pm0.04}}$ & 83.6$_{\textcolor{gray}{\pm0.02}}$ & 37.8$_{\textcolor{gray}{\pm0.01}}$ & 25.8$_{\textcolor{gray}{\pm0.32}}$ & 59.5$_{\textcolor{gray}{\pm0.12}}$ & 61.3$_{\textcolor{gray}{\pm0.06}}$ \\      
            PromptFL\cite{guo2022promptfl} &
            89.7$_{\textcolor{gray}{\pm0.01}}$ & 67.6$_{\textcolor{gray}{\pm0.01}}$ & 88.5$_{\textcolor{gray}{\pm0.07}}$ & 42.9$_{\textcolor{gray}{\pm0.08}}$ & 48.1$_{\textcolor{gray}{\pm0.22}}$ & 63.0$_{\textcolor{gray}{\pm0.01}}$ & 66.1$_{\textcolor{gray}{\pm0.02}}$ \\   
            TPFL (ours) &
            90.6$_{\textcolor{gray}{\pm0.01}}$ & 68.9$_{\textcolor{gray}{\pm0.01}}$ & 89.1$_{\textcolor{gray}{\pm0.00}}$ & 43.0$_{\textcolor{gray}{\pm0.07}}$ & 54.3$_{\textcolor{gray}{\pm0.29}}$ & 63.4$_{\textcolor{gray}{\pm0.01}}$ & 65.9$_{\textcolor{gray}{\pm0.02}}$ \\   
            \hline
            \hline
            {\textbf{\textsc{ATPFL}}} (ours) & \textbf{91.3}$_{\textcolor{gray}{\pm0.01}}$ & \textbf{69.6}$_{\textcolor{gray}{\pm0.01}}$ & \textbf{89.5}$_{\textcolor{gray}{\pm0.01}}$ & \textbf{44.1}$_{\textcolor{gray}{\pm0.01}}$ & \textbf{54.9}$_{\textcolor{gray}{\pm0.26}}$ & \textbf{63.8}$_{\textcolor{gray}{\pm0.00}}$ & \textbf{66.5}$_{\textcolor{gray}{\pm0.01}}$\\
            \hline
     \end{tabular}}
    \vspace{-6pt}
    \label{result_vit}
\end{table*}

\begin{table*}[t]
    \caption{Test Accuracy (\%) Results for ResNet-50 model on 7 datasets with 5 different seeds.}
    \vspace{-8pt}
    \centering
    \resizebox{0.99\textwidth}{!}{
        \setlength\tabcolsep{5pt}
        \renewcommand\arraystretch{1.02}
        \begin{tabular}{c||c c c c c c c }
            \thickhline
            \rowcolor{mygray}
            Algorithm (RN50) & Caltect-101	& Flowers-102 & Oxford-Pets & DTD & EuroSAT & Stanford Car & UCF-101\\
            \hline
            \hline
            Local Training &
            63.1$_{\textcolor{gray}{\pm0.37}}$ & 18.7$_{\textcolor{gray}{\pm2.61}}$ & 30.8$_{\textcolor{gray}{\pm4.83}}$ & 22.5$_{\textcolor{gray}{\pm0.21}}$ & 19.2$_{\textcolor{gray}{\pm0.01}}$ & 20.1$_{\textcolor{gray}{\pm0.93}}$ & 34.3$_{\textcolor{gray}{\pm0.36}}$ \\      
            PromptFL\cite{guo2022promptfl} &
            84.8$_{\textcolor{gray}{\pm0.04}}$ & 58.7$_{\textcolor{gray}{\pm0.01}}$ & 85.3$_{\textcolor{gray}{\pm0.04}}$ & 35.7$_{\textcolor{gray}{\pm0.03}}$ & \textbf{33.4}$_{\textcolor{gray}{\pm0.03}}$ & 52.9$_{\textcolor{gray}{\pm0.02}}$ & 57.8$_{\textcolor{gray}{\pm0.07}}$ \\   
            TPFL (ours) &
            85.2$_{\textcolor{gray}{\pm0.02}}$ & 59.6$_{\textcolor{gray}{\pm0.01}}$ & \textbf{85.6}$_{\textcolor{gray}{\pm0.02}}$ & \textbf{37.4}$_{\textcolor{gray}{\pm0.03}}$ & 32.2$_{\textcolor{gray}{\pm0.04}}$ & 53.8$_{\textcolor{gray}{\pm0.01}}$ & 58.2$_{\textcolor{gray}{\pm0.03}}$ \\   
            \hline
            \hline
            {\textbf{\textsc{ATPFL}}} (ours) & \textbf{85.6}$_{\textcolor{gray}{\pm0.02}}$ & \textbf{60.5}$_{\textcolor{gray}{\pm0.00}}$ & 85.4$_{\textcolor{gray}{\pm0.04}}$ & 36.9$_{\textcolor{gray}{\pm0.03}}$ & 32.2$_{\textcolor{gray}{\pm0.01}}$ & \textbf{54.1}$_{\textcolor{gray}{\pm0.01}}$ & \textbf{59.3}$_{\textcolor{gray}{\pm0.04}}$\\
            \hline
    \end{tabular}}
    \vspace{-6pt}
    \label{result_rs}
\end{table*}

\section{Evaluation}
In this section, we perform intensive evaluations to verify the effectiveness of our proposed TPFL and ATPFL.

\subsection{Evaluation setup.}
\textbf{Few-shot Dataset  and  Data  partition.} 
Extended from PromptFL\cite{guo2022promptfl} who only evaluates their model on four datasets, we verify ATPFL in seven different datasets: Caltech-101 \cite{fei2004learning}, Oxford-Pets, Stanford Cars, OxfordFlowers-102, EuroSAT, UCF-101, and Describable Textures (DTD). Furthermore, in order to create the few-shot dataset, we set that each client has $n_k$ samples for each class. For the majority of our evaluations, we choose $n_k = 4$ meaning that each client has a four-shot dataset; besides, we investigate the effect of the shot size in the ablation section. For the non-IID setting in FL, we select the label-skewing method to emulate the heterogeneous local clients. 

\textbf{Models}
Following the existing work, we choose the ResNet-50 (RN50) and Visual Transformer model (ViT) as the backbone of the visual encoder, and the Transformer model as the textual encoder.

\textbf{Baselines.}
In our evaluation, we compare ATPFL with the following baselines: (1) Local training, where all clients train their own models in an offline manner, and no model transmission is conducted; (2) PromptFL, using only the textual modality; (3) TPFL, employing both the textual and visual modalities, but no InfoNCE loss. 

\textbf{Implementation details.}
To prevent the influence of randomness and ensure the fairness of our evaluations, each experiment setting has been performed in three identical random seeds, and then we average the results to get the final result. We use the Adam optimizer with learning rate $\alpha = 1e-3$, and the Cosine scheduler with $epoch=20$. Furthermore, 
For the implementation environment, we conduct our code on Python version 3.11.0 and Pytorch 1.13.0. We also use 4 NTX NVIDIA A6000 GPUs to run our code.

\subsection{Main results}
In this section, the experimental outcomes are assessed. Table 1 and Table 2 present the average test accuracy for ViT and RN50 backbones across seven diverse datasets in a non-IID setting. Both PromptFL and ATPFL consistently surpass local training, with margins extending up to 18.5\%. This is intuitive, as local training or full-model fine-tuning may lead to catastrophic forgetting. This issue is exacerbated by client data heterogeneity. These compounded factors significantly impede fine-tuning performance in the federated learning context, necessitating the exploration of PromptFL and ATPFL. For ViT, TPFL excels over PromptFL in six of the seven datasets, with margins spanning 0.1\% - 6.2\%, except for UCF-101 where TPFL lags by 0.2\%. When factoring in the standard error of test accuracy across multiple experiments, TPFL's advancements over previous methods are noticeable. Despite TPFL's success, limitations persist, leading to the proposal of ATPFL to better address these issues. Our ATPFL model outperforms the baseline by 0.4\% - 1.1\% across all datasets, illustrating ATPFL's potential to mitigate data heterogeneity in prompt federated learning scenarios.

In the ResNet-50 tests, TPFL outperforms local training and PromptFL in six of the seven datasets, except for the EuroAT dataset. Our ATPFL continues to surpass TPFL in four of the seven datasets, except for Oxford-Pets and DTD where ATPFL trails TPFL by 0.2\% and 0.5\% respectively. This could be due to the model disparities between ViT and ResNet-50.

In conclusion, our proposed ATPFL, leveraging the concept of contrastive learning, offers superior performance in handling data heterogeneity. These results corroborate our prior discussions in the methodology section.

\subsection{Ablation study}
In this section, we examine various factors influencing our model's performance, including the application of InfoNCE loss, number of shot size, and client quantity. 

\textbf{InfoNCE loss.} First off, we investigates the impact of InfoNCE loss (i.e., the difference between TPFL and ATPFL). As illustrated in \cref{result_vit} and \cref{result_rs}, ATPFL shows a clear advantage compared to TPFL. In 11 out of 14 experiments, ATPFL outperforms TPFL by a margin of up to 1.1\%.

\begin{figure}[htbp]
    \centering
    \includegraphics[width=0.94\linewidth]{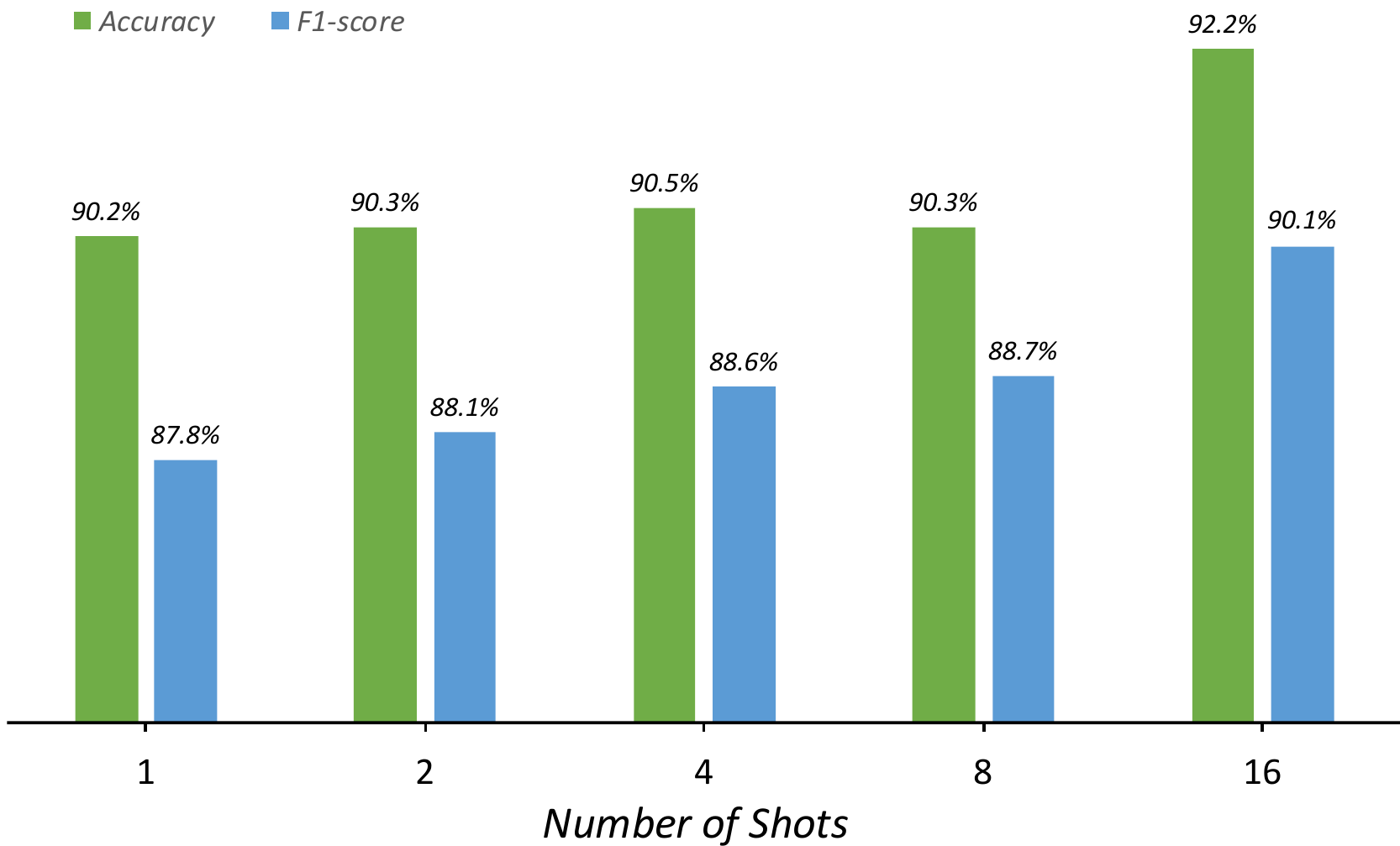}
    \vspace{-10pt}
    \caption{This figure illustrates how shot number affects the model accuracy and F1-score }
    \label{num_of_shots}
    \vspace{-15pt}
\end{figure}

\textbf{Shot size.} Second, we explore the impact of shot size, and \Cref{num_of_shots} demonstrates a monotonic increase in the F1-score as the number of shots rises, with the F1-score in a 16-shot scenario exceeding that of a 1-shot scenario by 2.3\%. Moreover, despite the absence of a consistent increase, accuracy still trends upward with an increasing number of shots. Even at a 1-shot scenario, ATPFL exhibits substantial performance (90.2\% accuracy and 87.8\% F1-score), but greater shot numbers offer additional potential performance benefits due to the increased feature information provided at each learning round.

\begin{figure}[htbp]
    \centering
    \includegraphics[width=0.94\linewidth]{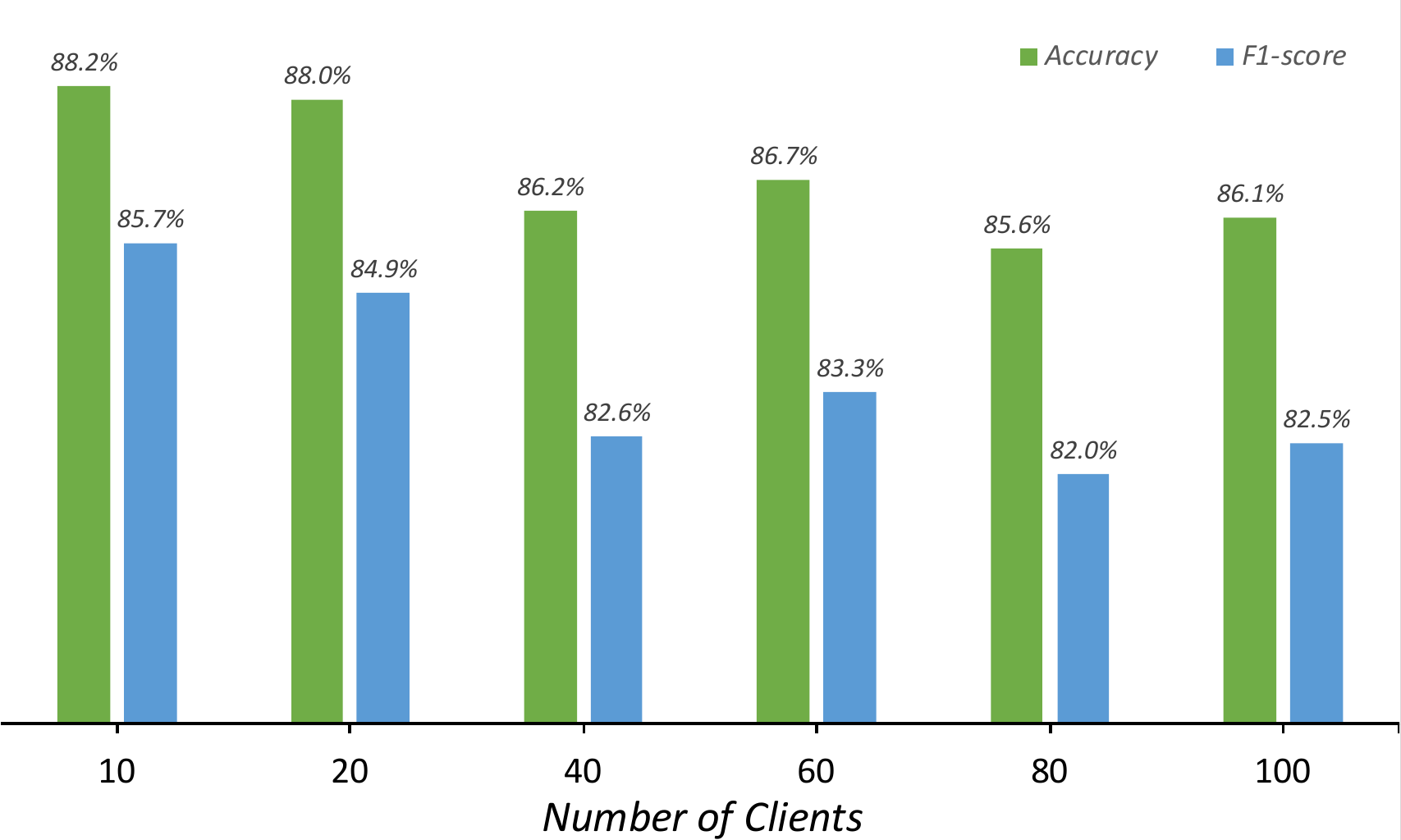}
    \vspace{-10pt}
    \caption{This figure illustrates how client number affects the model accuracy and F1-score}
    \label{num_of_clients}
    \vspace{-15pt}
\end{figure}

\textbf{Client volume.} Lastly, the ablation study examines the effect of the number of clients. \Cref{num_of_clients} reveals a decline in both accuracy and the F1-score as the client number rises, with a tenfold increase in clients (from 10 to 100) decreasing accuracy and the F1-score by 2.1\% and 3.2\%, respectively. However, even with a larger number of clients, ATPFL maintains reasonable performance, achieving 86.1\% accuracy in a 100-client scenario. 

\section{Conclusion}
In this paper, we propose an FL framework, TPFL, which first considers both visual and textual information in prompt tuning to augment the global model in FL. Notwithstanding, the performance improvement offered by TPFL is limited due to data heterogeneity. To address this issue, we developed ATPFL to facilitate local clients in obtaining more information from the global model, thereby enhancing their representing performance. A series of experiments have been conducted to validate the effectiveness of our methods, demonstrating that ATPFL consistently outperforms all baseline methods across various datasets and scenarios.

\begin{acks}
    This work was supported by the National Key R\&D Program of China under Grant No.2022ZD0160504, by Tsinghua Shenzhen International Graduate School-Shenzhen Pengrui Young Faculty Program of Shenzhen Pengrui Foundation (No. SZPR2023005), and by Tsinghua-Toyota Joint Research Institute inter-disciplinary Program and Tsinghua University (AIR)-Asiainfo Technologies (China) Inc. Joint Research Center under grant No. 20203910074. We would also like to thank anonymous reviewers for their insightful comments.
\end{acks}
\bibliographystyle{ACM-Reference-Format}
\bibliography{sample-base}

\end{document}